%% file: report.tex
\documentclass{article}
\usepackage{times}
\usepackage{algorithm}
\usepackage{algorithmic}
\usepackage{graphicx} 
\usepackage{subfigure} 
\usepackage{microtype}
\usepackage[sort&compress]{natbib}
\usepackage{algorithm}
\usepackage{algorithmic}
\usepackage{amsmath}
\usepackage{amssymb}
\usepackage{flushend}
\usepackage{booktabs}
\usepackage{multirow}
\usepackage{booktabs}
\usepackage{todonotes}
\usepackage{hyperref}
\usepackage[accepted]{icml2014ed} 
\begin{document}
\twocolumn[
\icmltitle{Recurrent Spatial Transformer Networks}
\icmlauthor{S\o ren Kaae S\o nderby$^\text{1}$}{skaaesonderby@gmail.com}
\icmlauthor{Casper Kaae S\o nderby$^\text{1}$}{casperkaae@gmail.com}
\icmlauthor{Lars Maal\o e$^\text{2}$}{larsma@dtu.dk}
\icmlauthor{Ole Winther$^\text{1,2}$}{olwi@dtu.dk}
\icmladdress{
$^\text{1}$ Bioinformatics Centre, Department of Biology, University of Copenhagen, Copenhagen, Denmark\\ 
$^\text{2}$ Department for Applied Mathematics and Computer Science, Technical University of Denmark, 2800 Lyngby, Denmark}

\icmlkeywords{Transformer network}
\icmltitlerunning{Transformer network}
\vskip 0.3in
]
\begin{abstract}
We integrate the recently proposed spatial transformer network (SPN) \citep{Jaderberg2015} into a recurrent neural network (RNN) to form an RNN-SPN model. We use the RNN-SPN to classify digits in cluttered MNIST sequences. The proposed model achieves a single digit error of 1.5\% compared to 2.9\% for a convolutional networks and 2.0\% for convolutional networks with SPN layers. The SPN outputs a zoomed, rotated and skewed version of the input image. We investigate different down-sampling factors (ratio of pixel in input and output) for the SPN and show that the RNN-SPN model is able to down-sample the input images without deteriorating performance. The down-sampling in RNN-SPN can be thought of as adaptive down-sampling that minimizes the information loss in the regions of interest. We attribute the superior performance of the RNN-SPN to the fact that it can attend to a sequence of regions of interest. 
\end{abstract}%
%
\input{introduction.tex}
\input{spn.tex}
\input{experiments.tex}
\bibliographystyle{icml2015}
\bibliography{library.bib}
\clearpage
\input{appendix.tex}

\end{document}

%% file: introduction.tex
\setcounter{footnote}{0}
\section{Introduction} 
\begin{figure*}[htbp]
	\centering
	\includegraphics[width=0.75\textwidth]{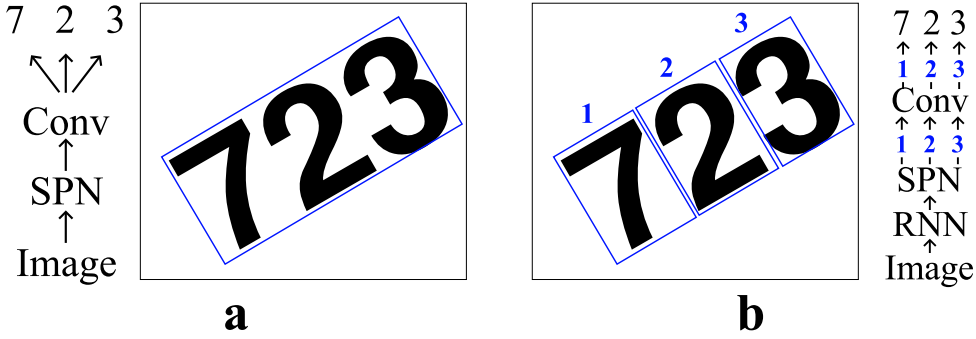}
	\caption{SPN networks for predicting the sequence 723 from an image.
	a) A FFN-SPN network attend to the entire sequence (blue box). The digits are classified by three separate softmax layers at the top of the network. Because the network cannot zoom in on individual digits in the sequence all digits are classified from the same image crop. b) A RNN-SPN where the transformation is predicted with an RNN. This allows the model to create a separate crop for each digit. Each crop (indicated with blue numbers) is then passed through the same classification network. The structure enables the model to zoom in on each individual digit.}
	\label{fig:ffnvsrnn}
\end{figure*}

Attention mechanisms have been used for machine translation \citep{Bahdanau2014},  
speech recognition \citep{Chorowski2014} and image recognition \citep{Xu2015,gregor2015draw,Ba2014}. The recently proposed spatial transformer network (SPN) \citep{Jaderberg2015} is a new method for incorporating spatial attention in neural networks. SPN uses a learned affine transformation of the input and bilinear interpolation to produce its output. This allows the SPN network to zoom, rotate and skew the input. A SPN layer can be used as any other layer in a feed-forward convolutional network\footnote{See e.g. \url{https://goo.gl/1Ho1M6}}. 
The feed-forward SPN (FFN-SPN) is illustrated in Figure~\ref{fig:ffnvsrnn} panel a) where a SPN combined with a convolutional network is used to predict a sequence of digits. Because the predictions are made with $n$ independent softmax layers at the top of the network, the SPN layers must find the box containing the entire sequence. 
A better model would sequentially produce the targets by locating and zoom in on each individual element before each prediction. In this paper we combine the SPN network with a recurrent neural network (RNN) to create a recurrent SPN (RNN-SPN) that sequentially zooms in on each element. This is illustrated in Figure~\ref{fig:ffnvsrnn} panel b). In the SPN-RNN a RNN network produces inputs for the SPN at each time-step. Using these inputs the SPN produces a transformed version of the input image which is then used for classification. By running the recursion for multiple steps and conditioning each transformation on the previous time-step the RNN-SPN model can sequentially attend to the part of the image containing each elements of interest and only use the relevant information for classification. Because these regions are generally small we experiment with forcing the RNN-SPN to down-sample the image which can thought of as adaptive down-sampling thus keeps the resolution of the regions of interest (nearly) constant. 

\section{Related Work}
\citealt{gregor2015draw} introduced a differentiable attention mechanism based on an array of 
Gaussians and combined it with a RNN for both generative and discriminative tasks. \citealt{Ba2014} and 
\citealt{Sermanet} combined a non-differentiable attention mechanism with an RNN and used it for classification. Their attention mechanism was trained using reinforcement learning.
Other related work include \cite{Xu2015} who applies visual attention in an encoder-decoder structure. 

%% file: spn.tex
\section{Spatial Transformer Network}%
\begin{figure*}[tb]
	\centering
	\includegraphics[width=0.75\textwidth]{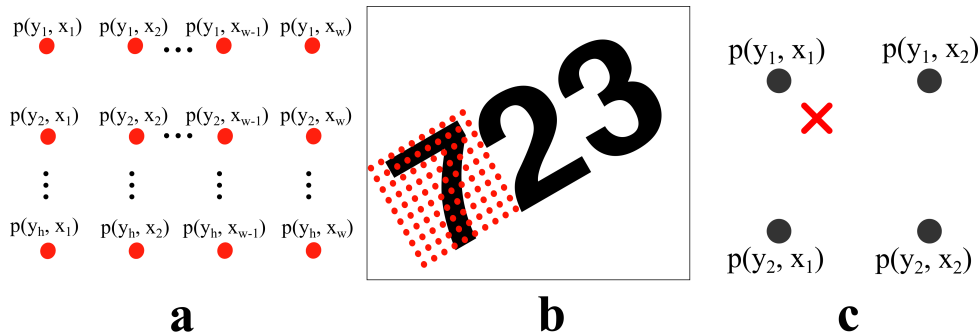}
	\caption{ a) The sampling grid $\mathbf{G}$ of equally spaced sampling points. We set 
	$y_1$ and $x_1$ to $-1$ and $y_h$, $x_w$ to $+1$. b) A recurrent SPN is able to zoom in on each element in the sequence. c) Bilinear transformation will interpolate the red cross by calculating a weighted average of the four nearest pixels. The operation is differentiable. 
}
	\label{fig:sampling}
\end{figure*}

The SPN network is implemented similarly to \citep{Jaderberg2015}. A SPN network takes an image or a feature map from a convolutional network as input. An affine transformation and bilinear interpolation is then applied to the input to produce the output of the SPN. The affine transformation allows zoom, rotation and skew of the input. The parameters of the transformation are predicted using a localization network $f_\text{loc}$:

\begin{equation}
f_{\text{loc}}(I) = \mathbf{A}_\theta= \begin{bmatrix} \theta_1 & \theta_2 & \theta_3 \\ 
                                    \theta_4 & \theta_5 & \theta_6\end{bmatrix},
\end{equation}
where $I$ is the input to the SPN with shape $[H \times W \times C]$ (height, width, channels) and matrix $\mathbf{A}_\theta$ specifies the affine transformation. The affine transformation is applied on a mesh grid $\mathbf{G} \in \mathcal{R}^{h \times w}$:
\begin{align}
	\mathbf{G} =& \{ (y_1, x_1), (y_1, x_2), ...(y_2, x_1), (y_2, x_2), \nonumber\\
				&... (y_w, x_{h-1}), (y_h, x_w)\} \label{eq:grid}.
\end{align}
where $w$ and $h$ does not need to be equal to $H$ and $W$. $\mathbf{G}$ is illustrated in Figure~\ref{fig:sampling} panel a). The grid is laid out such that $(y_1, x_1) = (-1, -1)$ and $(y_h, x_w) = (+1, +1)$ with the points in-between spaced equally. The affine transformation is applied to $\mathbf{G}$ to produce an image $\mathbf{S}$ which tells how to select points from $I$ and map them back onto $\mathbf{G}$:

\begin{align}
	\mathbf{S}_{ij} &= \mathbf{A}_\theta  \begin{bmatrix}y_i \\ x_j \\ 1 \end{bmatrix} 
\end{align}
where we have augmented each point with a $1$. Since the mapped points in $\mathbf{S}$ does not correspond exactly to one pixel in $I$ bilinear interpolation is used to interpolate each point in $\mathbf{S}$. The sub-gradients for the bilinear interpolation are defined and we can use standard backpropagation through the transformation to learn the parameters in $f_\text{loc}$. The sampling process is illustrated in Figure~\ref{fig:sampling} where panel a) illustrates $\mathbf{G}$ and panel b) illustrates $\mathbf{S}$ after we have applied the transformation. 

\subsection{Down-sampling}%
We can vary the number of sampled points by varying $h$ and $w$. Having $h$ and $w$ less than $H$ and $W$ will downsample the input to the SPN. We specify the down sampling with $d$. $d$ larger than 1 will downsample the input. The number of sampled points form $I$ is:
\begin{equation}
	n_{points} = \left(\frac{H}{d}\right)\cdot \left(\frac{W}{d}\right) = \frac{H\cdot W}{d^2}.
\end{equation}
For 2D images the sampled points decrease quadratically with $d$.

\subsection{RNN-SPN}%
In the original FFN-SPN the localization network is a feed-forward convolutional neural network. We modify this model by letting an RNN predict the transformation matrices such that
\begin{align}
c &= f_\text{conv}(I) \\ 
h_t &= f_\text{loc}^\text{rnn}(c, h_{t-1}) \\
\mathbf{A}_\theta &= g(h_t).
\end{align}
where $f_\text{conv}$ is a convolutional network taking $I$ as input and creating a feature map $c$, $f_\text{loc}^\text{rnn}$ is an RNN, and $g$ is a FFN. Here an affine transformation is produced at each time-step from the hidden state of the RNN. Importantly the affine transformations are conditioned on the previous transformations through the time dependency of the RNN.

%% file: experiments.tex
\section{Experiments} 
We test the model on a dataset of sequences of MNIST digits cluttered with noise. The dataset was created by placing 3 random MNIST digits on a canvas of size 100 $\times$ 100 pixels. The first digits was placed by randomly sampling an $y$ position on the canvas. The x positions were randomly sampled subject to the entire sequence must fit the canvas and the digits are non-overlapping. Subsequent digits are placed by following a slope sampled from $\pm 45^\circ$. Finally the images are cluttered by randomly placing 8 patches of size $9 \times 9$ pixels sampled from the original MNIST digits. For the test, validation and training sets we sample from the corresponding set in the original MNIST dataset. We create 60000 examples for training, 10000 for validation, and 10000 for testing\footnote{The script for generating the dataset is available along with the rest of the code.}. Figure~\ref{fig:zooms} shows examples of the generated sequences.

As a baseline model we trained a FNN-SPN with the SPN layer following immediately after the input. The classification network had 4 layers of conv-maxpool-dropout layers followed by a fully connected layer with 400 units and finally a separate softmax layer for each position in the sequence. The convolutional layers had 96 filters with size $3 \times 3$ and rectified linear units were used for nonlinearity in both the convolutional and fully connected layers. For comparison we further train a purely convolutional network similar to the classification network used in the FFN-SPN.

The RNN-SPN use a gated recurrent unit (GRU) \citep{Chung2014} with 256 units. The GRU is run for 3 time steps. At each time step the GRU unit use $c$ as input. We apply a linear layer to convert $h_t$ into $\mathbf{A}_\theta^t$. The RNN-SPN is followed by a classification convolutional network similar to the network used in the FFN-SPN model, except that the convolutional layers only have 32 filters.

In all experiments, the localization networks had 3 layers of max-pooling convolutional layers. All convolutional layers had 20 filters with size $[3 \times 3]$. All models were trained with RMSprop \citep{Tieleman2012} down-sampling factors and dropout rates optimized on the validation set. A complete description of the models can be found in the Appendix.

The models were implemented using Theano \citep{Bastien2012} and Lasagne \citep{sander_dieleman_2015_27878}. The SPN has been merged into the Lasagne library\footnote{Available here: \url{http://goo.gl/kgSk0t}}. Code for models and dataset is released at \url{https://goo.gl/RspkZy}.
\begin{table}[t]
\centering
\caption{Per digit error rates on MNIST sequence dataset, d is the down-sampling factor.}\vspace{0.3cm}
\label{tab:results}
\begin{tabular}{ll}
\hline
\multicolumn{2}{l}{{\bf Cluttered MNIST Sequences}} \\
{\it Model}      & {\it Err. (\%)}           \\ \hline
RNN-SPN d=1      & 1.8                   \\
RNN-SPN d=2      & {\bf 1.5}              \\
RNN-SPN d=3      & 1.8                 \\
RNN-SPN d=4      & 2.3                 \\
FFN-SPN d=1      & 4.4                      \\
FFN-SPN d=2      & 2.0                       \\
FFN-SPN d=3      & 2.9                     \\
FFN-SPN d=4      & 5.3                      \\
Conv. net.       & 2.9                       \\ \hline
\end{tabular}
\end{table}

\section{Results}
\begin{figure*}[th]
	\centering
	\includegraphics[width=0.60\textwidth]{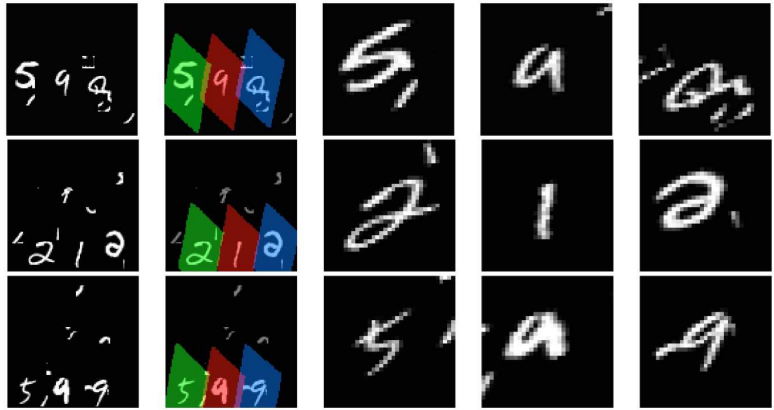}
	\caption{ The left column shows three examples of the generated cluttered MNIST sequences. The 
	next column shows where the model attend when classifying each digit. In the last 3 column we 
	show the image crops that the RNN-SPN uses to classify each digit. The input sequences are $100\times 100$ pixels. Each image crop is $33 \times 33$ pixels because the model uses a down-sample factor
	of 3.}
	\label{fig:zooms}
\end{figure*}

Table~\ref{tab:results} reports the per digit error rates for the tested models. The RNN-SPN models perform better than both convolutional networks (2.9\%) and FFN-SPN networks (2.0\%). In Figure~\ref{fig:zooms} we show where the model attend on three sample sequences from the test set. The last three columns show image crops after the affine transformation using a down-sampling factor of three. We found that increasing the down-sampling factor above one encouraged the model to zoom. When the down-sampling factor is greater than one we introduce an information bottleneck forcing the model to zoom in on each digit. The poor performance of the FFN-SPN convolutional net for high down-sampling values is explained by the effective decrease in resolution since the model needs to fit all three digits in the image crop.

\section{Conclusion}
We have shown that the SPN can be combined with an RNN to classify sequences. Combining RNN and SPN creates a model that performs better than FFN-SPN for classifying sequences. The RNN-SPN model is able to attend to each individual element in a sequence, something that the FFN-SPN network cannot do. The main advantage of the RNN-SPN model when compared to the DRAW network \citep{gregor2015draw} is that the SPN attention is faster to train. Compared with the model of \citep{Mnih2014} our model is end-to-end trainable with backpropagation.
In this work we have implemented a simple RNN-SPN model future work include allowing multiple glimpses per digit and using the current glimpse as input to the RNN network.

%% file: appendix.tex
\twocolumn[\icmltitle{Appendix}]
\subsection{RNN-SPN}
Localisation network
\begin{enumerate}\setlength\itemsep{-0.1cm}\footnotesize
  \item maxpool(2,2)
  \item conv(20@(3,3))
  \item maxpool(2,2)
  \item conv(20@(3,3))
  \item maxpool(2,2)
  \item conv(20@(3,3))
  \item GRU(units=256)
  \item Denselayer(6, linear output)
\end{enumerate}
Classification network
\begin{enumerate}\setlength\itemsep{-0.1cm}\footnotesize
	\item SPATIAL TRANSFORMER LAYER (downsample = {1, 3, 4})
    \item conv(32@(3,3))
    \item maxpool(2,2)
    \item dropout(p)
    \item conv(32@(3,3))
    \item maxpool(2,2)
    \item dropout(p)
    \item conv(32@(3,3))
    \item dropout(p)
    \item Dense(256)
    \item softmax
\end{enumerate}
In table Table~\ref{tab:results} the model are reported at RNN SPN with an entry for
each tested downsample factor.

\subsection{Baseline models}
\textbf{FFN-SPN model}
\begin{enumerate}\setlength\itemsep{-0.1cm}\footnotesize
	\item SPATIAL TRANSFORMER LAYER (downsample = 2.0)
    \item conv(96@(3,3))
    \item maxpool(2,2)
    \item dropout(p)
    \item conv(96@(3,3))
    \item maxpool(2,2)
    \item dropout(p)
    \item conv(96@(3,3))
    \item dropout(p)
    \item Dense(400)
    \item 3 $\times$ softmax
\end{enumerate}

Localisation network
\begin{enumerate}\setlength\itemsep{-0.1cm}\footnotesize
  \item maxpool(2,2)
  \item conv(20@(3,3))
  \item maxpool(2,2)
  \item conv(20@(3,3))
  \item maxpool(2,2)
  \item conv(20@(3,3))
  \item Denselayer(200)
  \item Denselayer(6, linear output)
\end{enumerate}
This model is reported in Table~\ref{tab:results} as the SPN conv. network.
Finally we also trained the classification network from the FFN-SPN reported as conv. Net 
in Table~\ref{tab:results}.